\documentclass[runningheads]{llncs}

\usepackage[final,year=2026]{eccv}
\usepackage{eccvabbrv}
\usepackage{graphicx}
\usepackage{booktabs}
\usepackage{tabularx,array}
\usepackage{amsmath,amssymb}
\usepackage[breaklinks,colorlinks,citecolor=eccvblue]{hyperref}

\graphicspath{{figures/}{overleaf_p3/figures/}}

\newif\ifincludesupplement
\includesupplementfalse 

\begin{document}

\title{Lensless Gaze Is Not Private by Default: Auditing Identity Leakage Across Disclosure Surfaces}
\titlerunning{Auditing Identity Leakage in Simulated Lensless Gaze}

\author{Rahul Vimalkanth\inst{} \and
Kaushik Mitra\inst{}}
\authorrunning{Vimalkanth \& Mitra}
\institute{Indian Institute of Technology Madras, Chennai, India\\
\email{ee23b061@smail.iitm.ac.in,\;kmitra@ee.iitm.ac.in}}

\maketitle

\begingroup
\renewcommand{\thefootnote}{}
\footnotetext{\textbf{Code:} \href{https://github.com/xoxo121/Lensless-Gaze-Is-Not-Private-by-Default}{github.com/xoxo121/Lensless-Gaze-Is-Not-Private-by-Default}}
\addtocounter{footnote}{-1}
\endgroup

\begin{abstract}
Lensless near-eye sensing is often described as privacy-friendly because its coded measurements are visually unintelligible. Yet visual unintelligibility reflects human interpretation, not what a learned adversary can recover. We therefore treat identity privacy as a systems property of disclosure surfaces: representations crossing sensing, storage, computation, and output boundaries. We audit a simulated lensless gaze pipeline under a common \(36\)-subject known-gallery closed-set identification protocol. We study a fixed, known-PSF setting in which the same optical encoding is used across enrollment and evaluation; privacy from an unknown or varying optical key is outside our scope. Reported accuracies are empirical attack success rates under matched linear and multilayer perceptron (MLP) probes; they demonstrate achievable leakage and do not upper-bound stronger adversaries.
Simulated lensless measurements yield \(96.7\%\) top-1 identification accuracy versus \(97.7\%\) for matched original eye crops, while a masked autoencoder (MAE) embedding retains \(94.3\%\). Compression alone offers little protection: an \(8\)-D principal component analysis (PCA) projection retains \(93.2\%\), and a matched \(8\)-D bottleneck on the same frozen MAE backbone retains \(91.8\%\), whereas separately trained \(8\)-D Gaze Semantic Projection Latent (GSPL) bottlenecks yield \(77.5\%\) mean recovery across three training seeds. A released \(128\)-way gaze token lowers single-frame recovery to \(38.1\%\), while its residual and continuous gaze output expose \(62.1\%\) and \(72.6\%\), respectively. Under a source-frame-disjoint tiled protocol, token summaries reach \(39.9\%\) at \(T{=}25\), showing that repeated-output risk depends on representation and aggregation.
These rates reflect all subject-correlated information in the evaluated dataset, including acquisition and behavioral cues, rather than isolating intrinsic ocular biometrics. Ordinary least squares residualization against a six-dimensional crop geometry and intensity summary still leaves lensless recovery at \(95.1\%\). Our results show that privacy claims for lensless sensing must be tested at disclosure boundaries rather than inferred from appearance.
\end{abstract}

\section{Introduction}
\label{sec:intro}

Gaze sensing supports interaction, accessibility, extended reality, and behavioral interfaces, but its pipelines process sensitive subject-linked information~\cite{kroger2020privacy,zhang2017its,wood2014eyetracking}. Eye appearance, periocular geometry, acquisition characteristics, and gaze behavior can all support identity inference~\cite{bednarik2012eye,kinnunen2010eye}. Lensless imaging is attractive for near-eye sensing because coded optics can replace bulky lenses and produce measurements that do not resemble conventional eye photographs~\cite{asif2017flatcam,huang2013lensless}. This visual unintelligibility can encourage an unsafe inference: if a human cannot recognize the eye, the measurement is private.

That inference conflates human interpretability, reconstruction quality, and resistance to machine inference. A visually unintelligible measurement may still preserve stable subject-correlated structure accessible to a learned attacker~\cite{fredrikson2015model,song2017privacy,rigaki2024survey}. Conversely, a compact semantic output may reveal little appearance while exposing behavioral identity. Privacy therefore cannot be assigned to an optical front end in isolation.

We instantiate this audit in a simulated lensless gaze pipeline derived from the Open Eye Dataset (OpenEDS)~\cite{OpenEDS2019}. A known-gallery attacker is enrolled with labeled samples from the same \(36\) identities encountered during evaluation and performs \(36\)-way closed-set identification. The gaze model is trained on subjects disjoint from this gallery. Each disclosure surface is evaluated independently using matched linear and MLP probes on held-out temporal blocks. Our setting is intentionally distinct from optical-encryption systems such as OpEnCam~\cite{khan2024opencam}: OpEnCam treats camera-specific optical elements as a secret key and evaluates attacks with partial or no knowledge of that key, whereas our stored PSF is fixed across captures and assumed known. We therefore ask what identity remains recoverable when optical-key secrecy or diversity is not itself the privacy mechanism.

The leakage ladder is non-monotonic. Original crops and simulated lensless measurements yield \(97.7\%\) and \(96.7\%\) top-1 identification, while a masked-autoencoder embedding retains \(94.3\%\). An \(8\)-D PCA projection retains \(93.2\%\), a matched \(8\)-D bottleneck on the same frozen MAE backbone \(91.8\%\), and separately trained \(8\)-D GSPL bottlenecks \(77.5\%\) mean recovery across three seeds, showing that dimensionality alone does not explain leakage reduction. At the output boundary, a \(128\)-way gaze token reduces single-frame recovery to \(38.1\%\), while its residual and continuous gaze output expose \(62.1\%\) and \(72.6\%\). Under a source-frame-disjoint tiled protocol, token summaries reach \(39.9\%\) at \(T{=}25\), showing that repeated-release risk depends on representation and aggregation.

The audit intentionally measures all subject-correlated information available in the dataset. A six-dimensional geometry and intensity baseline reaches \(95.5\%\), indicating that positioning, illumination, crop geometry, and related acquisition cues contribute strongly. Residualizing the flattened lensless measurement against these cues on enrollment frames only reduces recovery from \(96.7\%\) to \(95.1\%\). Such cues remain privacy-relevant when they cross a trust boundary, but limit causal interpretation: the experiments establish empirical recoverability under the evaluated dataset and protocol, not intrinsic ocular biometrics or cross-session persistence.

The contributions are:
\begin{enumerate}
\item \textbf{Disclosure-surface auditing.} We evaluate measured, stored, represented, retained, released, and temporally aggregated signals under one explicit attacker protocol.
\item \textbf{Controlled simulated-lensless audit.} We show near-parity between paired original crops and lensless measurements under matched preprocessing, acquisition-cue residualization, and full-resolution spatial probes, while separating recoverability from biometric causation and formal privacy.
\item \textbf{Compression controls.} A \(192\)-D embedding, its \(8\)-D PCA projection, and a matched \(8\)-D bottleneck remain highly identifying, whereas separately trained low-dimensional GSPL bottlenecks leak less across multiple training seeds.
\item \textbf{Output- and time-aware analysis.} We audit a released gaze token, locally retained residual under GazeSplit, continuous prediction, behavioral controls, and repeated-output summaries.
\end{enumerate}

\begin{figure}[t]
\centering
\includegraphics[width=\linewidth]{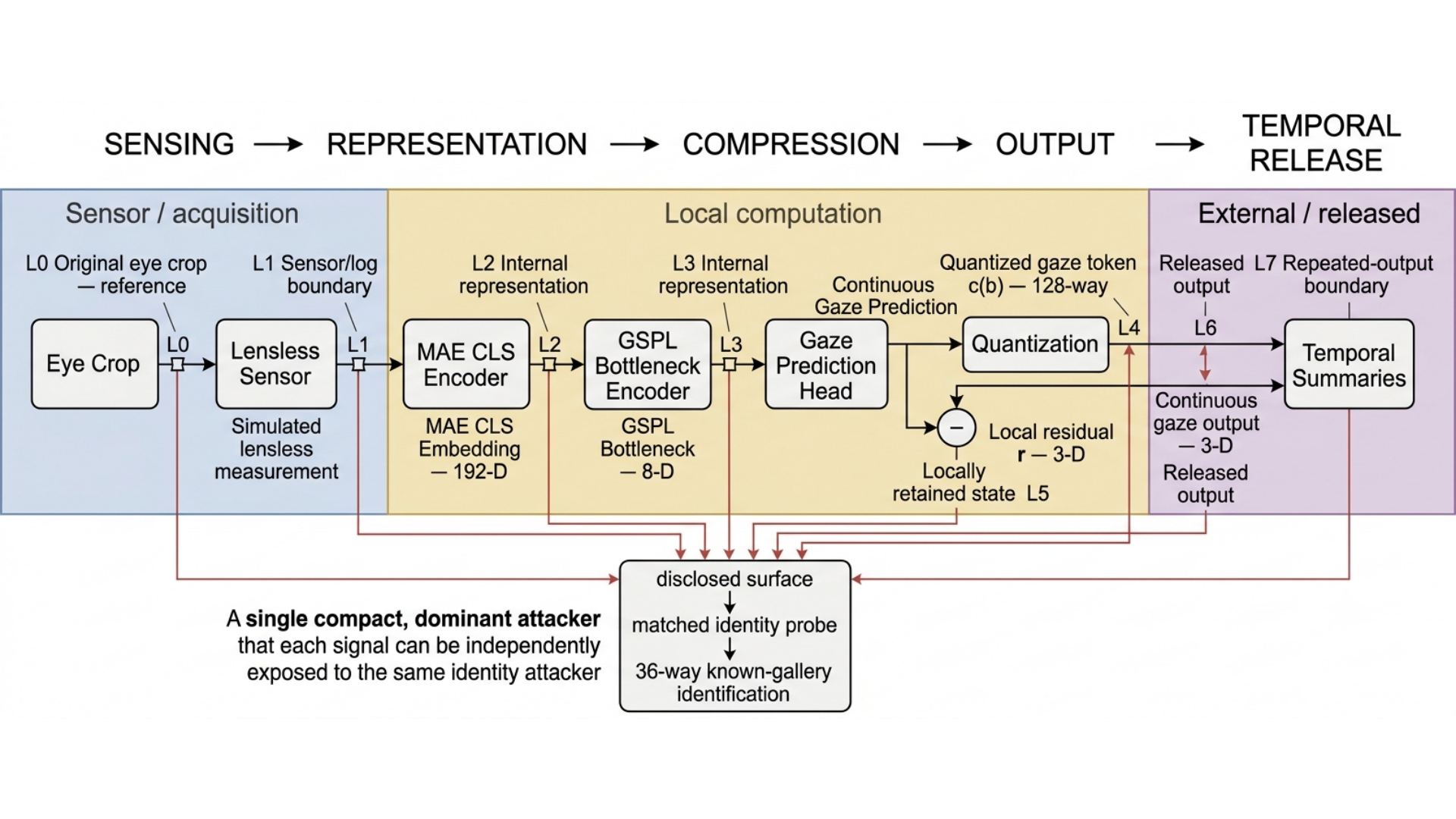}
\caption{Disclosure surfaces spanning sensing, internal representations, gaze outputs, and repeated release. Each surface is independently exposed to the same known-gallery identity attacker.}
\label{fig:overview}
\end{figure}

\section{Related Work}
\label{sec:related}

\paragraph{Identity Leakage in Eye Tracking.}
Eye tracking is widely used in augmented and virtual reality, accessibility, and human--computer interaction~\cite{zhang2017its,wood2014eyetracking}. Eye data encode appearance, physiology, behavior, and cognitive state, making them inherently privacy-sensitive~\cite{kroger2020privacy}. Eye images, periocular appearance, and gaze dynamics have all supported biometric recognition~\cite{bednarik2012eye,kinnunen2010eye,padole2011periocular}; recent work further shows identification from eye-movement trajectories alone~\cite{lohr2022eyeknow,aziz2023crossplatform}. These studies motivate evaluating leakage from both image representations and gaze outputs.

\paragraph{Privacy-Preserving Gaze Systems.}
Recent work has explored privacy-preserving gaze acquisition, representation, and release. Differentially private mechanisms account for temporal correlations~\cite{bozkir2021temporal}, while privacy-preserving streaming and dataset-publication mechanisms study privacy--utility trade-offs for released gaze data~\cite{davidjohn2021streaming,davidjohn2023datasets}. PrivateGaze transforms facial appearance before gaze estimation~\cite{du2024privategaze}, and recent benchmarks evaluate iris obfuscation against biometric leakage and gaze utility~\cite{wang2025irisobfuscation}. Unlike these protection methods, we do not propose a defense; we audit leakage throughout the sensing pipeline.

\paragraph{Lensless imaging and task-oriented inference.}
Lensless cameras replace conventional optics with coded elements while preserving information for reconstruction or downstream inference~\cite{asif2017flatcam,antipa2018diffusercam,boominathan2020phlatcam,boominathan2022recent}. FlatTrack~\cite{jain2025flattrack} demonstrates gaze estimation with an ultra-thin lensless near-eye camera, while LenslessFace~\cite{cai2024lenslessface} performs identity verification directly from lensless measurements. OpEnCam~\cite{khan2024opencam} instead designs lensless optics explicitly for encryption: its multiplexing PSF and scaling mask form a camera-specific optical key, and its attack model assumes partial or no knowledge of that key. This differs from our fixed, known-PSF setting. Accordingly, our results should not be interpreted as contradicting privacy obtained from optical-key secrecy or diversity; we study identity leakage when the same known optical encoding is reused across enrollment and evaluation.

\paragraph{Representation leakage.}
Learned representations frequently retain information beyond their intended target. Model inversion, membership inference, and classifier probes are established tools for measuring such leakage~\cite{fredrikson2015model,shokri2017membership,song2017privacy,rigaki2024survey,alain2017understanding}. Rather than introducing a new probe, we apply a common enrolled attacker across the trust boundaries of one sensing pipeline: raw measurements, learned embeddings, compressed representations, local state, released outputs, and temporal aggregates. The goal is diagnostic empirical recoverability, not differential privacy, anonymity, or cryptographic confidentiality.

\section{Threat Model and Method}
\label{sec:threat}

This work audits subject-predictive information already present in measured and learned signals; it does not introduce privacy-preserving training.

\subsection{Threat Model and Two-Level Split}
\label{sec:threat-model}

The protected attribute is subject identity; angular gaze error is reported separately as utility. A known-gallery attacker is enrolled with labeled samples from the same identities encountered at evaluation. Each experiment reveals exactly one disclosure surface. The probes are matched empirical attackers, not optimal adversaries. \cref{tab:threat_model} summarizes the assumptions.

\begin{table}[t]
  \caption{Threat-model parameters for the disclosure-surface audit.}
  \label{tab:threat_model}
  \centering
  \small
  \setlength{\tabcolsep}{4pt}
  \renewcommand{\arraystretch}{0.96}
  \begin{tabularx}{\linewidth}{@{}lX@{}}
  \toprule
  Parameter & Assumption / configuration \\
  \midrule
  Gallery & \(36\) known enrolled identities \\
  Enrollment & \(75\) labeled frames/identity (\(2{,}700\) samples) \\
  Evaluation & \(25\) held-out frames/identity (\(900\) samples) \\
  Gaze model & Trained on subject-disjoint data \\
  Architecture / weights & Known / not provided to attacker \\
  PSF & Fixed across captures; stored PSF assumed known \\
  Representation & Definition and parsing rules known \\
  Attacker & Fixed linear and two-layer MLP probes \\
  Goal & \(36\)-way closed-set identification (chance \(2.8\%\)) \\
  \bottomrule
  \end{tabularx}
\end{table}

The gaze model uses a subject-disjoint 288/36/36 train/validation/test split; identity attackers operate only on the 36 held-out test subjects. For each subject, numerically ordered frames are partitioned into four consecutive 25-frame blocks; three enroll the attacker and one evaluates it, reducing direct leakage from adjacent frames and shared blink states.

Five probe seeds vary MLP initialization and the internal \(10\%\) training/validation subdivision used for early stopping while preserving the gallery and grouped enrollment/evaluation partition. The same probe family and hyperparameter procedure are used across surfaces, so differences measure recoverability under a matched attacker family, although dimensionality and representation-specific preprocessing can affect attack difficulty. The protocol does not evaluate open-set recognition, unseen identities, enrollment-free matching, or cross-session persistence.

\subsection{Disclosure Surfaces}
\label{sec:surfaces}

For disclosed representation \(s\), identity \(i\), probe \(\phi\), and held-out set \(\mathcal{D}_{\mathrm{eval}}\), empirical identity recoverability is
\begin{equation}
\operatorname{Leak}_{\phi}(s)
=
\operatorname{Acc}_{\mathrm{ID}}
\bigl(\phi(s),i;\mathcal{D}_{\mathrm{eval}}\bigr).
\label{eq:leakage}
\end{equation}
It is conditional on the attacker, gallery, split, and disclosure definition; it is neither mutual information nor a formal privacy bound.

\begin{table}[t]
  \caption{Evaluated disclosure surfaces and deployment boundaries. For categorical L4, \(128\)-way denotes vocabulary cardinality.}
  \label{tab:surfaces}
  \centering
  \small
  \setlength{\tabcolsep}{4pt}
  \renewcommand{\arraystretch}{0.94}
  \begin{tabular}{llcl}
  \toprule
  ID & Surface & Size / dim. & Exposure boundary \\
  \midrule
  L0 & Original eye crop & \(3072\) & Reference \\
  L1 & Simulated lensless measurement & \(3072\) & Sensor / log \\
  L2 & MAE CLS embedding & \(192\) & Internal \\
  L3 & GSPL bottleneck & \(8\) & Internal \\
  L3$^\prime$ & PCA of L2 & \(8\) & Internal control \\
  L4 & Quantized gaze token & \(128\)-way & Released output \\
  L5 & Local residual & \(3\) & Local state \\
  L6 & Continuous gaze output & \(3\) & Released output \\
  L7 & Temporal summaries & Varies & Repeated output \\
  \bottomrule
  \end{tabular}
\end{table}

\subsection{GazeSplit: Auditing a Public--Private Output Boundary}
\label{sec:gazesplit}

We define GazeSplit as an audit construct partitioning a continuous gaze estimate into a released quantized token and locally retained residual. Let \(\hat{\mathbf{g}}\in\mathbb{R}^{3}\) be a unit-norm gaze prediction. We convert it to yaw and pitch and quantize it with a fixed \(8\times16=128\)-bin vocabulary over pitch \([-30^\circ,30^\circ]\) and yaw \([-45^\circ,45^\circ]\). Token \(b\) is represented by bin-center direction \(\mathbf{c}(b)\in\mathbb{R}^{3}\). Only 28 of 128 bins are occupied on the test split, with empirical entropy \(4.21\) bits.

\begin{align}
b &= \tau(\hat{\mathbf{g}}), &
\mathbf{r} &= \hat{\mathbf{g}}-\mathbf{c}(b), \nonumber\\
\hat{\mathbf{g}} &= \mathbf{c}(b)+\mathbf{r}.
\label{eq:gazesplit}
\end{align}

The Cartesian residual \(\mathbf{r}\in\mathbb{R}^{3}\) is omitted when the continuous prediction is replaced by the public bin center. It is audited as hypothetical locally retained state and is not used by a correction module in the evaluated MAE pipeline; GazeSplit is therefore an analysis device, not a deployed end-to-end system. L5 could in practice remain on device for calibration, error correction, or foveated rendering while only the token is transmitted. Cartesian differencing exactly reconstructs \(\hat{\mathbf{g}}\), although \(\mathbf{r}\) is neither unit norm nor an independent gaze direction. Since \(\hat{\mathbf{g}}=\mathbf{c}(b)+\mathbf{r}\) and the occupied vocabulary carries only \(4.21\) bits of entropy, L5 and L6 are near-informationally related; their MLP gap should be read with secondary probes.

\begin{figure}[t]
\centering
\begin{minipage}[t]{0.43\linewidth}
  \centering
  \includegraphics[width=\linewidth]{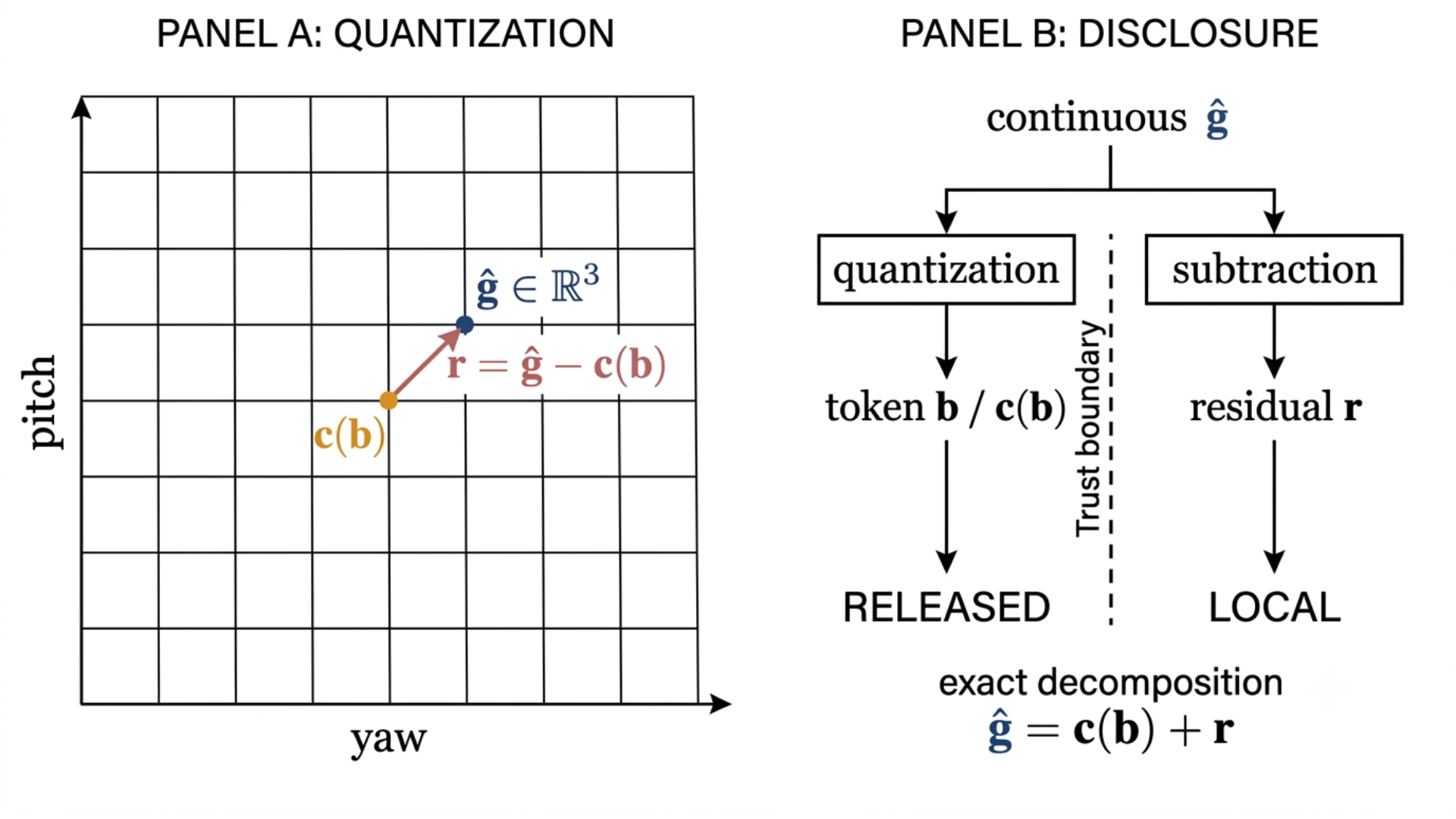}\\[-1mm]
  {\small\textbf{(a)} GazeSplit decomposition.}
\end{minipage}
\hfill
\begin{minipage}[t]{0.55\linewidth}
  \centering
  \includegraphics[width=\linewidth]{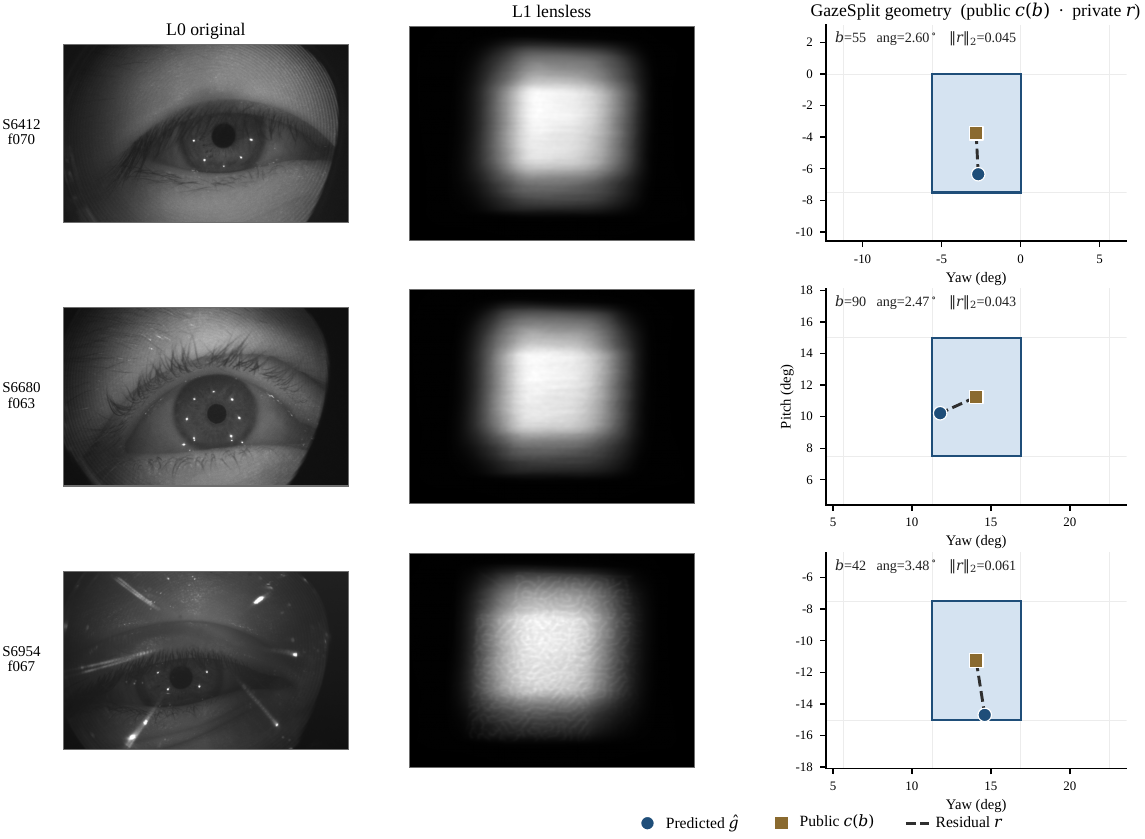}\\[-1mm]
  {\small\textbf{(b)} Representative held-out examples.}
\end{minipage}
\caption{\textbf{GazeSplit disclosure boundary.}
(a) A continuous prediction is decomposed into released token \(b\), represented by bin center \(\mathbf c(b)\), and locally retained residual \(\mathbf r\), with
\(\hat{\mathbf g}=\mathbf c(b)+\mathbf r\).
(b) Real held-out examples showing the corresponding decomposition; samples are not selected for identity-retrieval success.}
\label{fig:gazesplit}
\end{figure}

\section{Experimental Setup}
\label{sec:setup}

\begin{figure}[t]
  \centering
  \includegraphics[width=\linewidth]{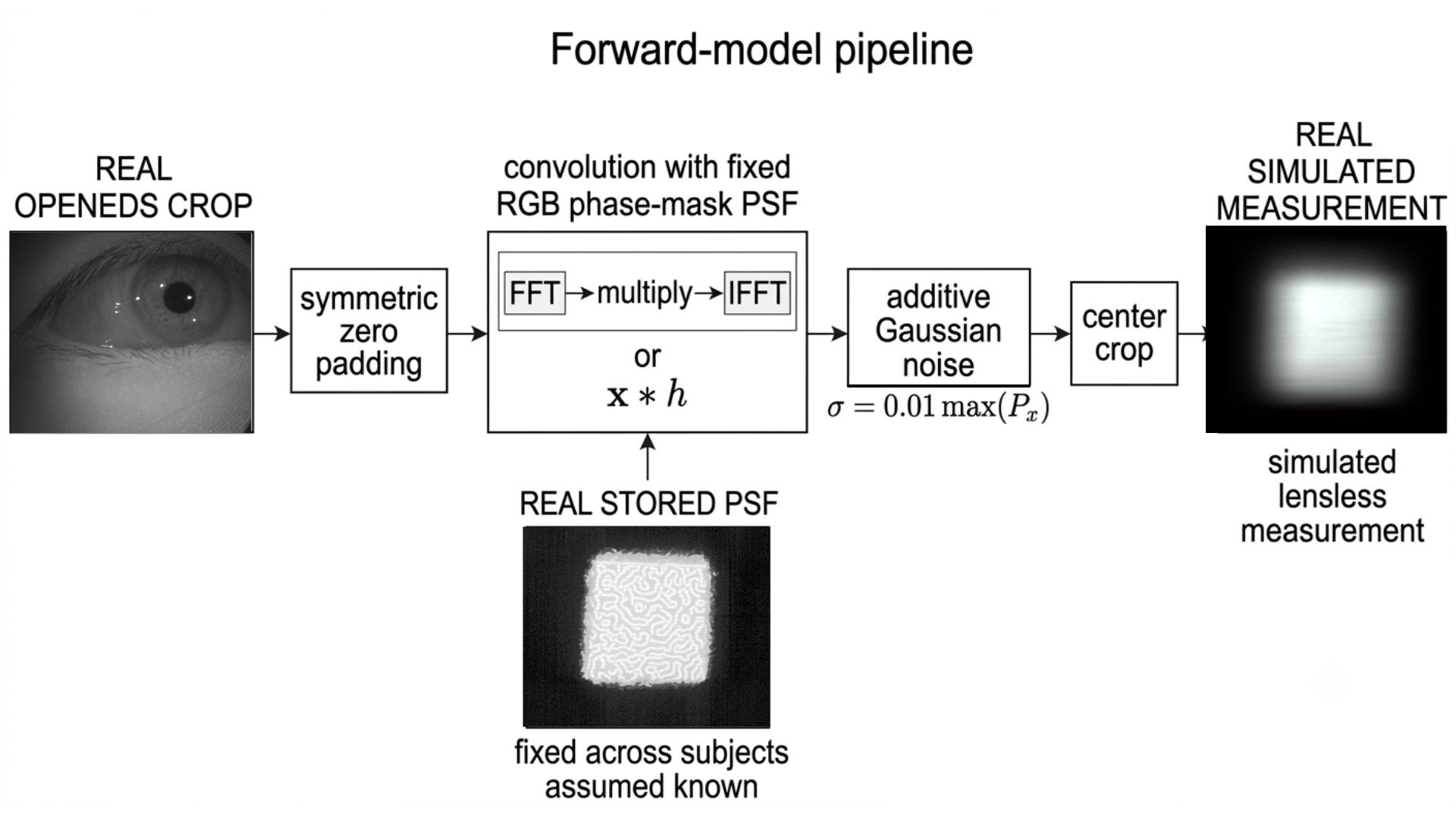}
  \caption{Forward simulation used to generate lensless measurements: OpenEDS crops are padded, convolved with the fixed stored RGB PSF, corrupted with Gaussian noise, and center-cropped.}
  \label{fig:simulation_triplet}
\end{figure}

\paragraph{Data and simulation.}
We derive simulated lensless measurements from OpenEDS. One fixed stored red--green--blue (RGB) phase-mask point-spread function (PSF) is used across all subjects and captures and is assumed known under our threat model, so we do not evaluate privacy arising from PSF secrecy or optical-key diversity. The fixed PSF also prevents identity from being encoded through subject-specific PSF assignment. The grayscale crop is replicated to three channels, symmetrically zero-padded, convolved with the RGB PSF using fast Fourier transforms (FFTs), corrupted by additive Gaussian noise \(\sigma=0.01\max(\mathbf{P}\mathbf{x})\), and center-cropped. \cref{fig:simulation_triplet} shows the corresponding forward-model pipeline.

L0 and L1 probes use matched preprocessing: resize to \(224\times224\), ImageNet channel normalization, bilinear downsampling to \(32\times32\), and flattening to \(3072\) dimensions. Thus the attacker receives matched preprocessed tensors rather than native-resolution files. As a spatial-resolution control, we extract frozen ImageNet-pretrained ResNet-18 features from the same \(224\times224\) frames and probe them under the identical grouped-block protocol.

\paragraph{Representations and controls.}
L2, L4, L5, and L6 use a lensless MAE with a Vision Transformer Tiny (ViT-Tiny) backbone and 192-D class-token (CLS) embedding. L3 is an \(8\)-D gaze bottleneck from a separately trained GSPL model predicting gaze without identity supervision; L2--L3 therefore mixes dimensionality, architecture, and objective. To isolate linear dimensionality reduction, L3$^\prime$ is an \(8\)-D PCA projection of L2 fitted per probe seed on attacker-training frames only. As a matched nonlinear control, we train an \(8\)-D gaze bottleneck on the same frozen MAE backbone, without identity supervision.

\paragraph{Attackers and uncertainty.}
We report matched linear and two-layer MLP probes. Primary-ladder intervals are descriptive percentile-bootstrap intervals over five probe-seed mean accuracies; they summarize probe-run variation only, not encoder retraining, subject resampling, or acquisition-session variation.

\section{Results}
\label{sec:results}

\subsection{Research Question 1 (RQ1): Does Visual Lensless Encoding Suppress Identity Recovery?}
\label{sec:results-raw}

The forward model removes recognizable ocular appearance from the displayed measurement (\cref{fig:simulation_triplet}), but visual change alone does not determine machine identity recovery.

Original crops (L0) yield \(97.7\%\) top-1 identification with descriptive interval \([96.7,98.5]\), and simulated lensless measurements (L1) \(96.7\%\) with \([95.4,98.4]\). Their paired mean difference is \(0.96\) percentage points, with descriptive interval \([-0.18,1.84]\). This is not an equivalence test or evidence of statistical equality; the reduction is small under the evaluated probe runs and matched preprocessing.

\paragraph{What is being recovered?}
The attack predicts subject labels from any stable subject-correlated information in the disclosed signal. A six-dimensional source-crop geometry/intensity summary, crop centroid plus channel means and variances, reaches \(95.5\%\). Thus high L0/L1 accuracy cannot be attributed solely to intrinsic ocular appearance; acquisition-linked geometry, positioning, illumination, and preprocessing regularities contribute strongly. These rates reflect all subject-correlated information available in the dataset, not only intrinsic ocular biometrics, and do not establish cross-session persistence.

To test whether removing these cues collapses recovery, we fit an ordinary least squares map from the six-dimensional summary to each coordinate of flattened L0/L1 using enrollment blocks only, then probe the residuals. Lensless recovery falls from \(96.7\%\) to \(95.1\%\), and original-crop recovery from \(97.7\%\) to \(96.2\%\). Concatenating cues with residualized L1 yields \(96.4\%\), only \(0.9\) points above cues alone; residualizing L2 similarly leaves \(93.8\%\) (from \(94.3\%\)). Linear residualization therefore does not remove identifying structure available to the matched attacker within session.

A frozen ImageNet-pretrained ResNet-18 probe on native \(224\times224\) frames yields \(97.5\%\) for original crops and \(95.8\%\) for lensless measurements, close to the flattened \(32\times32\) MLP results; near-parity is therefore not an artifact of aggressive downsampling alone. Grouped-block enrollment is also less optimistic than random splitting: on L1, random splits reach \(99.7\%\), grouped blocks \(96.7\%\), and max-gap blocks \(91.5\%\). Even the stricter within-sequence partition remains far above chance, but none creates independent headset placements or recording sessions.

A four-fold leave-one-block-out (LOBO) check yields mean MLP accuracies of \(99.2\%\) for L0 and \(98.4\%\) for L1. LOBO varies the held-out temporal block, not subject partition or session, supporting partition robustness of near-parity without addressing cross-session generality.

\begin{table}[tb]
  \caption{Linear identification for stored nominal L1 and matched forward-model resimulations. These test local simulation stability, not generality across hardware.}
  \label{tab:psf_robustness}
  \centering
  \small
  \setlength{\tabcolsep}{4pt}
  \renewcommand{\arraystretch}{0.94}
  \begin{tabular}{lc}
  \toprule
  PSF / simulation condition & Accuracy (\%) \\
  \midrule
  Stored nominal L1 TIFF & 97.2 \\
  Matched nominal resimulation & 95.3 \\
  Alternative phase-mask PSF & 96.6 \\
  Horizontal shift (2 pixels) & 95.7 \\
  Additive noise (\(0.01\max(\mathbf{P}\mathbf{x})\)) & 95.6 \\
  \bottomrule
  \end{tabular}
\end{table}

Matched forward-model resimulations yield \(95.3\%\)--\(96.6\%\) linear accuracy under the tested conditions (\cref{tab:psf_robustness}), versus \(97.2\%\) for stored nominal L1. Near-parity is therefore stable to these simulated variations. The \(96.7\%\) primary MLP rate nevertheless depends on the chosen simulation: a structurally different, highly diffusive phase mask might reduce recovery more substantially. These values demonstrate vulnerability of this stored-PSF simulation, not a universal leakage bound for physical coded optics.

\subsection{RQ2: Where Does Identity Persist Across the Pipeline?}
\label{sec:results-ladder}

\begin{figure}[t]
\centering
\begin{minipage}[t]{0.50\linewidth}
  \centering
  \includegraphics[width=\linewidth]{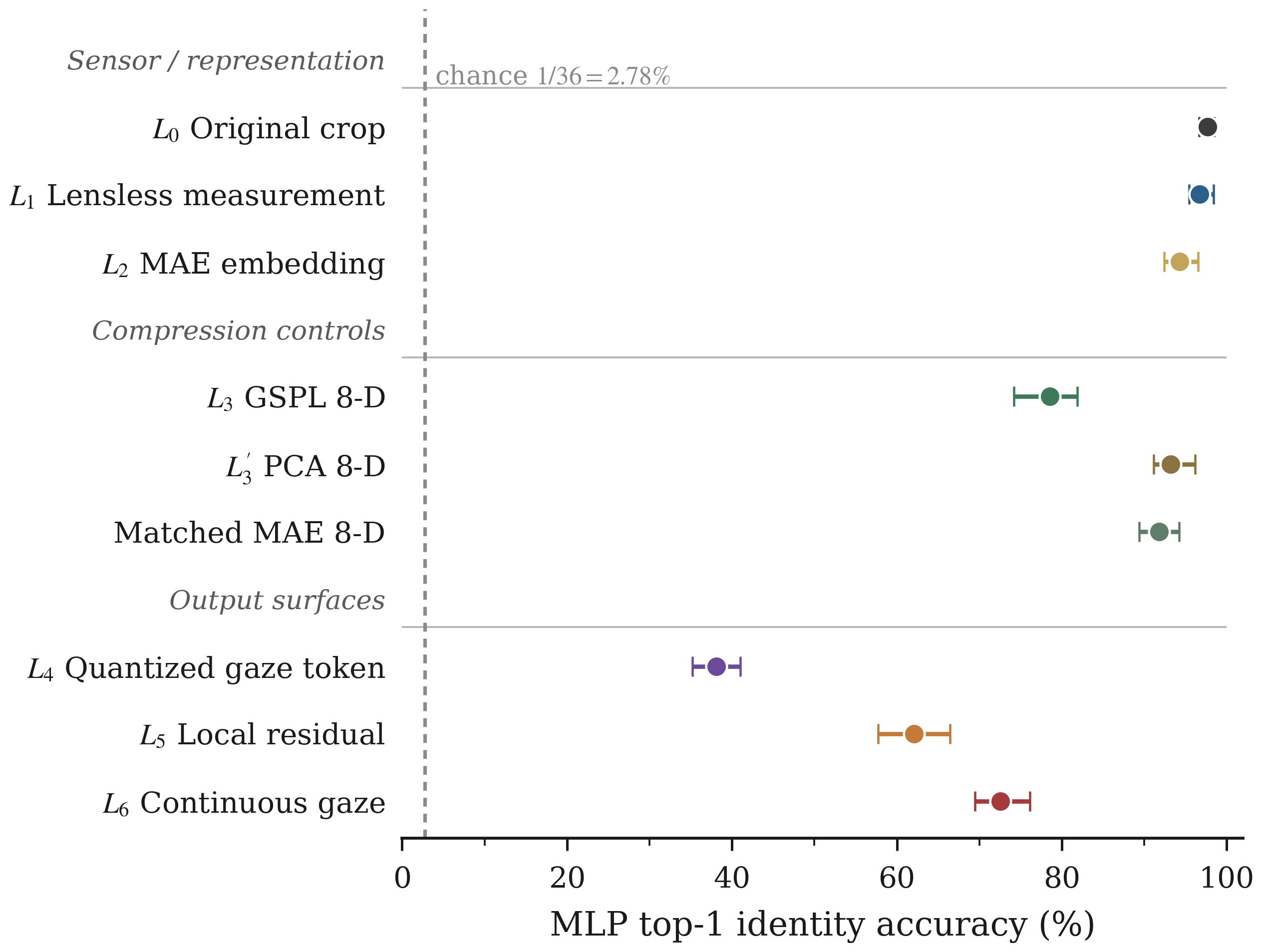}\\[-1mm]
  {\small\textbf{(a)} Leakage ladder.}
\end{minipage}\hfill
\begin{minipage}[t]{0.48\linewidth}
  \centering
  \includegraphics[width=\linewidth]{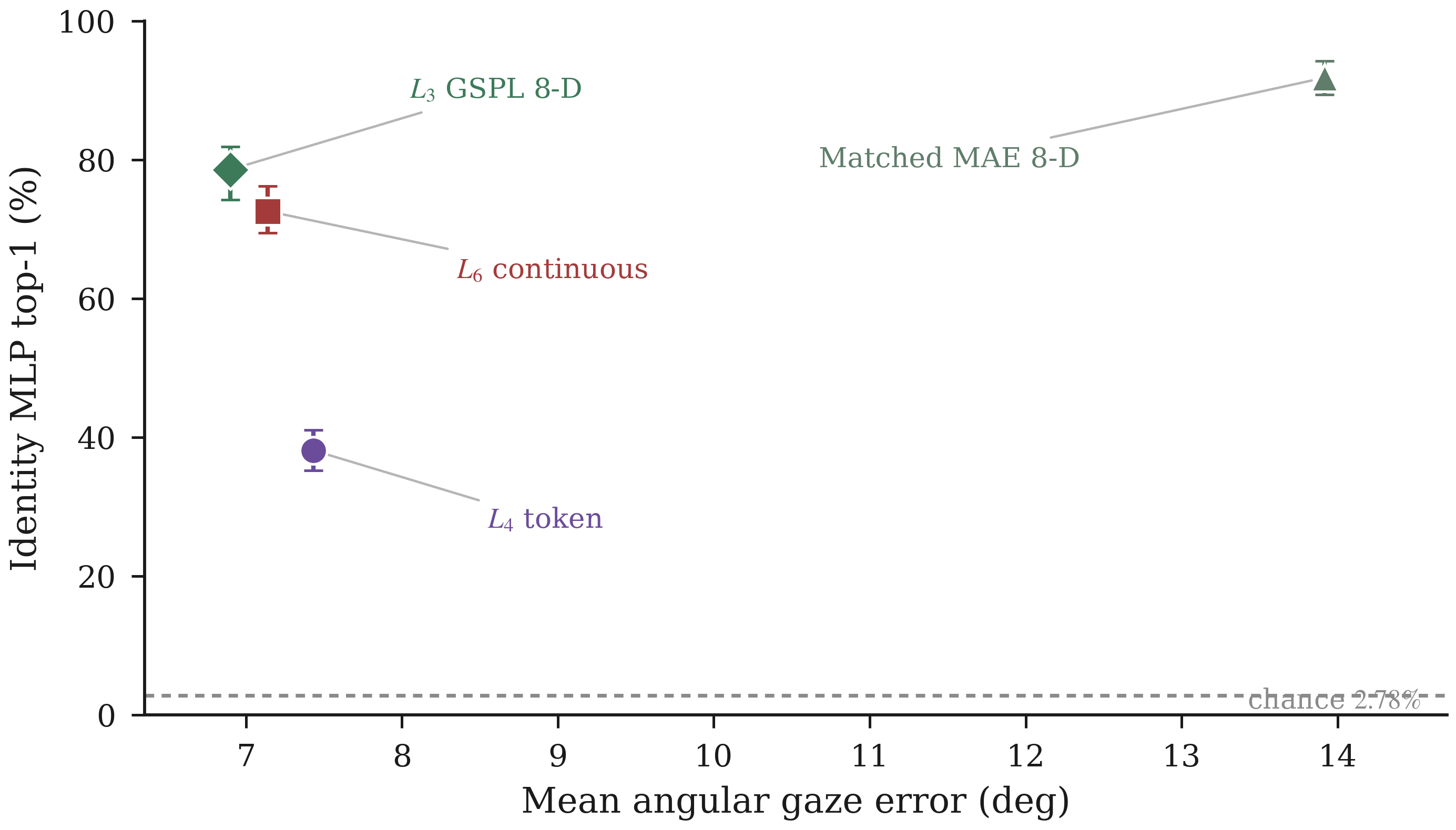}\\[-1mm]
  {\small\textbf{(b)} Privacy--utility comparison.}
\end{minipage}
\caption{\textbf{Identity leakage across representations and utility.}
(a) Grouped-block MLP top-1 identity recovery across the principal disclosure surfaces and compression controls, with the descriptive intervals reported in \cref{tab:ladder}; chance is \(1/36\).
(b) Identity recovery versus gaze error for representations with comparable utility measurements. Lower values on both axes are preferable; the matched \(8\)-D bottleneck is a compression control rather than a competitive gaze representation.}
\label{fig:main_results}
\end{figure}

\cref{tab:ladder,fig:main_results} summarize the leakage ladder. Recoverability does not decrease monotonically with dimensionality or apparent compression. L2 retains \(94.3\%\). L3 yields \(78.6\%\) for the primary run; three separately trained GSPL seeds yield \(78.6\%\), \(75.4\%\), and \(78.5\%\) (mean \(77.5\%\)), showing the reduction is not a single-run artifact. In contrast, L3$^\prime$ retains \(93.2\%\), and a matched \(8\)-D gaze bottleneck on the same frozen MAE backbone retains \(91.8\%\). Thus most linearly preserved subject-discriminative variation in L2 lies in its leading principal subspace, and a width-matched same-backbone bottleneck remains highly identifying. Low dimension alone is not a privacy mechanism; lower GSPL recovery indicates that representation and training choices materially affect leakage. The matched MAE bottleneck has worse gaze utility (\(13.9^\circ\) versus \(7.14^\circ\) for L2 and \(6.90^\circ\) for GSPL), so it is a compression control rather than a competitive gaze representation.

\begin{table}[tb]
  \caption{Primary grouped-block MLP leakage ladder. Intervals are descriptive percentile-bootstrap intervals over five probe-seed means and exclude encoder-retraining and subject-split uncertainty. \(128\) denotes categorical vocabulary cardinality.}
  \label{tab:ladder}
  \centering
  \small
  \setlength{\tabcolsep}{3.5pt}
  \renewcommand{\arraystretch}{0.92}
  \begin{tabular}{lccc}
  \toprule
  Disclosure surface / control & Size / dim. & Acc. (\%) & Interval \\
  \midrule
  \textit{Sensor and representation surfaces} & & & \\
  L0: Original eye crop & 3072 & 97.7 & [96.7, 98.5] \\
  L1: Simulated lensless measurement & 3072 & 96.7 & [95.4, 98.4] \\
  L2: MAE embedding & 192 & 94.3 & [92.4, 96.6] \\
  L3: GSPL bottleneck (separate model) & 8 & 78.6 & [74.2, 81.9] \\
  L3$^\prime$: PCA projection of L2 & 8 & 93.2 & [91.2, 96.2] \\
  Matched MAE \(8\)-D bottleneck & 8 & 91.8 & [89.4, 94.2] \\
  \midrule
  \textit{Gaze-output surfaces} & & & \\
  L4: Quantized gaze token & 128-way & 38.1 & [35.2, 41.0] \\
  L5: Local residual & 3 & 62.1 & [57.8, 66.4] \\
  L6: Continuous gaze output & 3 & 72.6 & [69.4, 76.2] \\
  \midrule
  \textit{Behavioral controls} & & & \\
  Ground-truth continuous gaze & 3 & 76.4 & [71.4, 81.3] \\
  Ground-truth quantized token & 128-way & 43.1 & [41.2, 45.1] \\
  Chance (\(36\) identities) & -- & 2.8 & [2.8, 2.8] \\
  \bottomrule
  \end{tabular}
\end{table}

The released token (L4) gives the lowest single-frame recovery, \(38.1\%\), but remains far above \(2.8\%\) chance. Continuous gaze (L6) reaches \(72.6\%\), and residual L5 \(62.1\%\). Because \(\hat{\mathbf{g}}=\mathbf{c}(b)+\mathbf{r}\), the L4 \(\ll\) L6 ordering partly follows from quantizing a continuous output to about \(4.2\) bits, whereas the residual is a less obvious disclosure surface a designer might log or cache. L5/L6 ordering is probe-dependent: under a linear probe L5/L6 reach
\(34.1/64.3\%\), while under \(5\)-nearest neighbors they reach
\(81.3/84.6\%\). The systems finding is that omitted local state can remain identifying, not that the MLP gap is intrinsic. Tokenization reduces empirical recovery, but privacy still depends on whether residual or full-precision predictions are exposed elsewhere.

For outputs of the same MAE ViT-Tiny model, L4, L5, and L6 yield \(38.1\%\), \(62.1\%\), and \(72.6\%\) identity recovery, respectively. Quantized L4 incurs \(7.43^\circ\) mean angular error versus \(7.14^\circ\) for continuous L6; angular gaze error is not applicable to residual L5. Thus the large reduction in identity recovery from L6 to L4 accompanies only a \(0.29^\circ\) increase in gaze error under this model (\cref{fig:main_results}).

Applying the same vocabulary to ground-truth labels yields \(43.1\%\) from quantized gaze and \(76.4\%\) from continuous gaze. Their proximity to predicted-gaze recovery indicates that subject-specific gaze distributions explain substantial output leakage, although prediction errors and model artifacts may contribute.

Interpreting continuous-gaze leakage nevertheless requires caution. Given the \(95.5\%\) geometry/intensity baseline, genuine behavioral gaze dynamics cannot be definitively disentangled from dataset-specific positioning or cropping artifacts; model predictions may partially proxy these static acquisition biases.

Laplace-smoothed \(128\)-bin token histograms yield mean base-2 Jensen--Shannon (JS) divergence \(0.368\) between subjects, \(3.89\times\) the first-half/second-half within-subject divergence. Subjects also occupy sparse token subsets: median \(2.0\) unique bins per subject among 28 occupied overall. Thus token identification within session may reflect which recording-specific bins were visited rather than a stable cross-session behavioral biometric. Without independent sessions, this establishes dataset-level behavioral separability, not durable identity across headset placements.

As an open-set-adjacent diagnostic, verification thresholds calibrated on 12 identities and evaluated on 24 disjoint identities yield calibration/evaluation equal-error rates of \(8.0/2.6\%\) for L2 and \(15.8/12.2\%\) for L6. This is not open-set identification and does not replace the closed-set threat model.

\subsection{RQ3: How Does the Release Protocol Change Risk?}
\label{sec:results-accumulation}

Single-frame accuracy does not characterize repeated releases. We form consecutive non-overlapping tiles ordered by frame index, using earliest windows for enrollment and latest for evaluation to ensure source-frame disjointness. Token windows use normalized \(128\)-bin histograms; continuous and residual windows are mean-pooled.

\begin{table}[tb]
  \caption{Identity recovery under source-frame-disjoint tiles; mean \(\pm\) standard deviation over \(20\) trials.}
  \label{tab:accumulation}
  \centering
  \small
  \setlength{\tabcolsep}{5pt}
  \renewcommand{\arraystretch}{0.94}
  \begin{tabular}{lcc}
  \toprule
  Surface & \(T=1\) & \(T=25\) \\
  \midrule
  Quantized gaze token & \(34.0 \pm 1.9\) & \(39.9 \pm 2.0\) \\
  Continuous unit gaze & \(55.7 \pm 3.0\) & \(41.8 \pm 0.6\) \\
  Local residual & \(44.9 \pm 2.1\) & \(36.8 \pm 1.9\) \\
  \bottomrule
  \end{tabular}
\end{table}

\begin{figure}[t]
  \centering
  \includegraphics[width=0.90\linewidth]{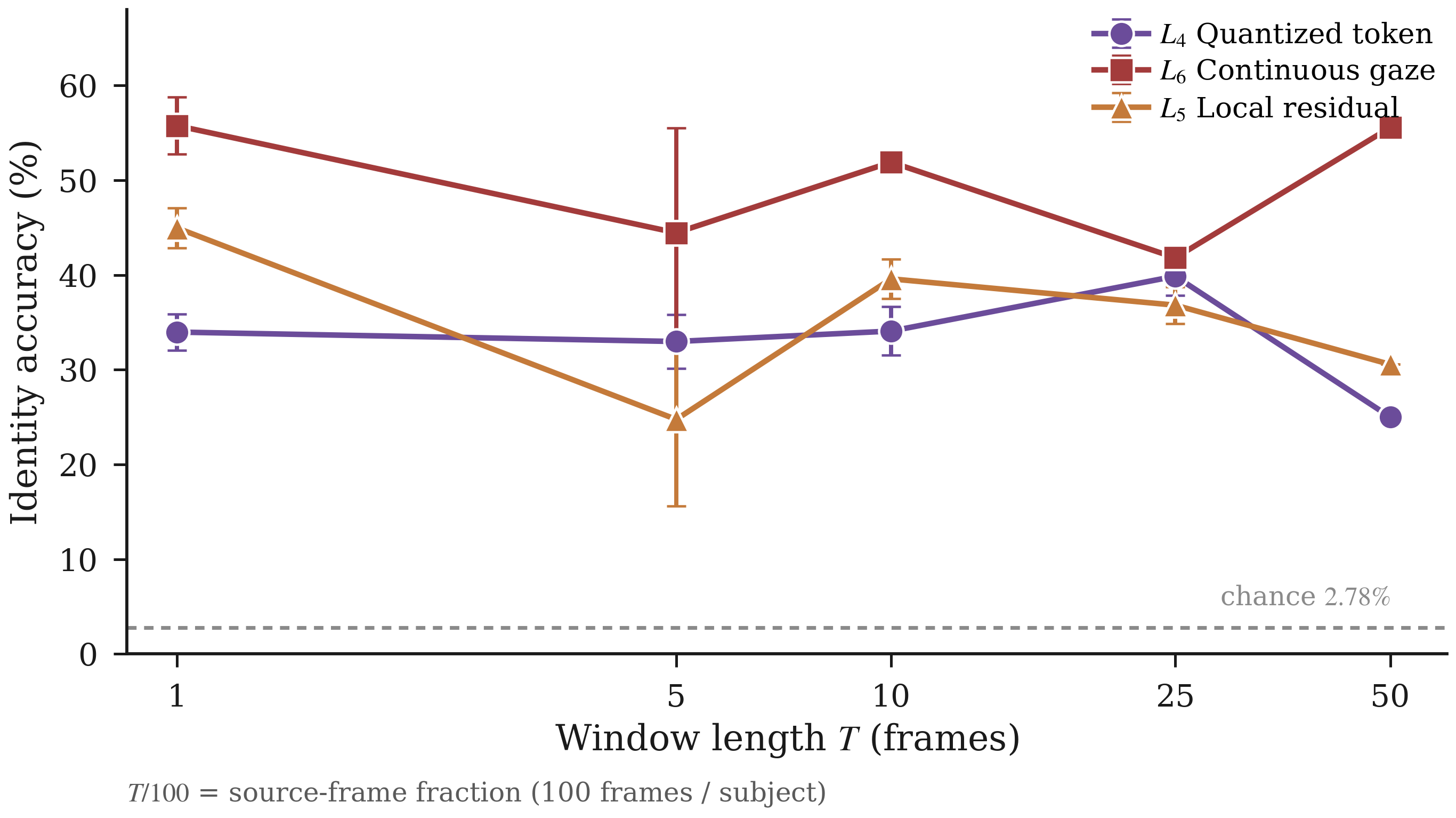}
  \caption{Identity recovery under source-frame-disjoint temporal aggregation. Token histograms and mean-pooled continuous/residual outputs exhibit different trends with release duration.}
  \label{fig:temporal_release}
\end{figure}

Token recovery rises from \(34.0\%\) at \(T{=}1\) to \(39.9\%\) at \(T{=}25\), whereas mean-pooled continuous and residual outputs decrease (\cref{fig:temporal_release}).

The \(T{=}1\) values differ from \cref{tab:ladder} because the tiled protocol requires a different enrollment/evaluation split to enforce source-frame disjointness across sequential windows. This protocol shift avoids train/test contamination and supports a narrow conclusion: temporal privacy risk depends on representation and aggregation. Histograms can expose stable token-frequency differences, while mean pooling can suppress frame-level cues. We do not claim monotonic leakage with window length or generality to alternative sequence models, pooling operators, or longer streams.

A separate ViT-Small calibration experiment reaches \(1.42^\circ\) gaze error when subject-specific calibration data and fitted parameters remain on device, showing useful local personalization is possible. It is not integrated with GazeSplit and does not establish a complete privacy--utility frontier.

\section{Discussion}
\label{sec:discussion}

\paragraph{What the audit establishes.}
Visually unintelligible simulated lensless measurements can remain highly subject-predictive under an enrolled attacker, including after acquisition-cue residualization and under a full-resolution spatial probe. Disclosure risk is non-monotonic: neither optical obfuscation nor low dimensionality guarantees reduced recoverability. Semantically minimal outputs can also expose subject-linked information when residuals or continuous predictions are retained or repeatedly released. Privacy-sensitive systems should therefore be audited at each trust boundary rather than judged from one representation's appearance.

A cue need not be an intrinsic biometric to constitute systems-level leakage: if sufficiently stable to identify an enrolled user and disclosed across a boundary, it is part of the attack surface. This does not establish persistence across sessions, devices, or headset placements; it establishes identifying disclosure under the evaluated conditions.

\paragraph{What the audit does not establish.}
The rates are simulation- and protocol-specific, not universal properties of lensless imaging. They do not isolate intrinsic ocular biometrics from acquisition cues, demonstrate cross-session persistence, evaluate unseen identities, or provide formal privacy guarantees. OpenEDS provides ordered frames within subject folders but no independent session identifiers, so enrollment and evaluation share acquisition conditions. The appropriate interpretation is subject-correlated recoverability under a specified disclosure and threat model, not an intrinsic information-theoretic identity bound.

\paragraph{Design implications.}
\begin{itemize}
    \item Audit the system, not the image: human-unrecognizable measurements may support machine identification.
    \item Treat stable geometry, positioning, illumination, behavior, and learned features crossing trust boundaries as attack surface.
    \item Do not equate bottlenecks with privacy: PCA-8D preserves \(93.2\%\) and the matched MAE \(8\)-D control \(91.8\%\); privacy requires explicit objectives and evaluation, not compression alone.
    \item Minimize and isolate outputs: tokenization reduces recovery, but residuals and continuous predictions remain sensitive if exposed.
    \item Audit release streams: risk depends on duration, aggregation, occupancy, and attacker adaptation, not only per-frame disclosure.
\end{itemize}

The disclosure-surface view generalizes beyond gaze: enumerate every signal crossing a trust boundary, define the attacker for that boundary, and measure the protected attribute directly. Optical coding may provide operational advantages, but these should be stated separately from identity privacy.

\section{Limitations and Ethical Considerations}
\label{sec:limitations}

\paragraph{Scope of evaluation.}
Results use a simulated lensless pipeline with one fixed optical configuration and evaluation protocol. This differs from optical-encryption settings such as OpEnCam~\cite{khan2024opencam}, where camera-specific optical elements form a secret key unavailable to the attacker. Our results therefore do not characterize identity leakage when the optical key is unknown or varies across cameras or captures. Such key uncertainty or diversity may reduce recoverability and is an important direction for future disclosure-surface audits. Perturbation experiments demonstrate robustness to moderate acquisition variations, but physical lensless hardware and structurally distinct optical designs also remain future work.

\paragraph{Threat-model coverage.}
Linear and MLP probes establish achievable leakage but do not upper-bound stronger adversaries, which may attain different absolute performance. The central conclusion is therefore limited to showing that privacy cannot be inferred solely from visual unintelligibility in the fixed, known-PSF setting evaluated here.

\paragraph{Generality.}
Evaluation uses one subject partition and a closed-set enrolled-attacker protocol. Additional subject partitions, independent sessions with distinct headset placements, and broader deployment conditions are needed to further characterize generality.

The work audits privacy risk rather than advocating biometric identification. We encourage privacy-by-design practices: minimizing retention of raw measurements and unrestricted embeddings, keeping calibration information local where possible, limiting unnecessary logging, and auditing learned representations before deployment.

\section{Conclusion}
\label{sec:conclusion}

We presented a disclosure-surface audit for identity leakage in a simulated lensless gaze pipeline. Under a common known-gallery threat model with a fixed, known PSF, we measured recoverability from sensor measurements, learned representations, compressed bottlenecks, locally retained state, released outputs, and temporal aggregates. Visually unintelligible lensless measurements remain highly identifying under this evaluated setting, including after linear residualization against acquisition cues; compression alone does not ensure privacy; and quantized gaze tokens reduce but do not eliminate leakage when residuals or continuous outputs remain available.

More broadly, privacy should be evaluated as a property of the entire sensing pipeline rather than optical encoding alone. The disclosure-surface perspective provides a practical framework for auditing where sensitive information persists and where effective privacy mechanisms are required.

\section*{Acknowledgements}
K.M. acknowledges funding support from the Anusandhan National Research Foundation (ANRF), India, under grant CRG/2023/007358, and the Qualcomm Faculty Award 2024.

\bibliographystyle{splncs04}
\bibliography{references}

\ifincludesupplement
\clearpage
\appendix
\input{supplement}
\fi

\end{document}